\documentclass{article}
\usepackage{microtype}
\usepackage{graphicx}
\usepackage{subcaption}
\usepackage{booktabs}
\usepackage{hyperref}

\usepackage[accepted]{icml2026}
\usepackage{amsmath}
\usepackage{amssymb}
\usepackage{mathtools}
\usepackage{amsthm}
\usepackage[capitalize,noabbrev]{cleveref}
\theoremstyle{plain}

\theoremstyle{definition}

\theoremstyle{remark}

\usepackage[textsize=tiny]{todonotes}
\icmltitlerunning{Beyond Accuracy: A Dual-Judge Protocol for Legally Grounded VLM Evaluation}
\begin{document}
\twocolumn[
\icmltitle{Beyond Accuracy: A Dual-Judge Evaluation Protocol\\
for Vision-Language Models in Legally Grounded Tasks}
\icmlsetsymbol{equal}{*}
\begin{icmlauthorlist}
\icmlauthor{Su Myat Noe}{nii}
\icmlauthor{Ha Thanh Nguyen}{vin}
\icmlauthor{May Myo Zin}{rois}
\icmlauthor{Ken Satoh}{rois}
\end{icmlauthorlist}
\icmlaffiliation{nii}{National Institute of Informatics (NII) / LLMC, Tokyo, Japan}
\icmlaffiliation{vin}{VinUniversity, Hanoi, Vietnam}
\icmlaffiliation{rois}{ROIS-DS Center for Juris-Informatics, Tokyo, Japan}
\icmlcorrespondingauthor{Su Myat Noe}{sumyatnoe@nii.ac.jp}
\icmlkeywords{AI for Law, LLM-as-judge, evaluation methodology,
              vision-language models, traffic-sign interpretation,
              legal reasoning}
\vskip 0.3in
]
\printAffiliationsAndNotice{}
\begin{abstract}
AI systems are increasingly evaluated for legally accountable settings,
where correct outputs must also be justifiable against an applicable
legal standard. Existing legal-AI benchmarks and LLM-as-judge protocols
provide important infrastructure for measuring task performance and
open-ended response quality. We contribute one additional evaluation
signal: a \emph{dual-judge protocol} that pairs a standard $0$--$10$
quality judge with a strict binary semantic-equivalence judge against a
human-curated reference. We study a controlled, visually grounded
regulatory task---UK traffic-sign interpretation, whose meaning is a
codified question with a known reference for every input---and measure not
merely \emph{whether} the two judges disagree (by construction they must)
but \emph{how much} and \emph{where}. On $4{,}680$ evaluations under seven
visibility levels and two occlusion modes, the two judges are moderately
associated (point-biserial $r=0.644$), while revealing an asymmetric
Type~II pattern affecting $8.0\%$ of all evaluations. Its distribution is
instructive: the \emph{marginal} rate peaks at high visibility ($14.2\%$ at
$v=0.8$) simply because high-scoring answers are common there, but
\emph{conditioned} on the answer already scoring above $7$ the rate is
highest under heavy occlusion ($54$--$63\%$ at $v\leq0.3$)---so a high
quality score is least trustworthy when the input is most degraded. We are
explicit that the signal is a property of this judge and reference: a $49$-row human check shows the $0$--$10$ judge
aligns closely with everyday-reader judgement (Pearson $r=0.81$; $r=0.80$
with the LLM accuracy sub-score) while the equivalence judge is fairly but
one-directionally stricter. The protocol adds one LLM call per evaluation
and surfaces a signal single-judge protocols do not report. We release the
prompt template, occluded variants, and full evaluation results.
\end{abstract}
\section{Introduction}
AI systems are increasingly deployed in domains where the law
requires decisions to be \emph{justifiable against an applicable
legal standard} alongside being correct: contract review, judicial
decision-support, regulatory compliance, and autonomous systems
operating under traffic, aviation, or medical law. The AI for Law
community has responded by building evaluation infrastructure for
this space, including LexGLUE~\cite{chalkidis2022lexglue},
LegalBench~\cite{guha2023legalbench}, and LLM-as-judge
methodology~\cite{zheng2023judging}.
\paragraph{What this paper contributes.}
We propose one additional signal that \emph{complements} this
infrastructure rather than replaces it. Concretely, we add a
strict binary \emph{semantic-equivalence judge} alongside the
standard quality judge: a yes/no LLM call asking only whether the
response is semantically equivalent to a human-curated reference.
We call the resulting two-score evaluation a
\emph{dual-judge protocol}.
The argument for the additional signal is straightforward. In
legally accountable deployment, two responses can score equally
well on an everyday-quality scale while only one would survive an
audit against the applicable reference. A single quality
judge cannot distinguish between them by design --- not because
the judge is wrong, but because the question it was asked is
different from the question an auditor asks. Adding a strict
equivalence judge gives an evaluator a second, complementary view
of the same response.
\paragraph{State-of-the-art gap.}
This dual-judge framing is, to our knowledge, not yet standard
practice in legal-AI evaluation. LexGLUE and LegalBench score
against gold labels or expert adjudications and report a single
accuracy-style metric; recent LLM-as-judge work
\cite{zheng2023judging} typically reports a single quality score.
The strict semantic-equivalence question --- ``does this response
match the reference?'' --- is not currently part of the
default protocol, so the disagreement between ``sounds right''
and ``matches the reference'' is not currently surfaced.
This paper offers a small, reproducible demonstration of what
that signal looks like in practice, on a domain where the
applicable rule is known for every input.
\paragraph{Testbed and scope.}
We use traffic-sign interpretation under the UK Road Traffic Act
and TSRGD as the testbed. Traffic regulations are codified law,
they are applied in real time by autonomous systems whose failures
have direct legal consequences, and the dataset structure permits
a controlled benchmark in which the ground-truth applicable rule
is known for every input.
We are deliberate about scope: this is a \emph{visually grounded
regulatory} task, and we use ``legal'' language throughout as
\emph{motivation} for why equivalence to a fixed reference matters, not
as a claim to evaluate open-textured legal \emph{interpretation}
(applying a rule to facts). Traffic-sign meaning is a codified question
with, in the ordinary case, a single correct reading; we return to this
distinction in Section~\ref{sec:discussion}.
Across $4{,}680$ evaluations of four vision-language
systems~\cite{openai2023gpt4,openai2024gpt4o} (single agent;
single agent with chain-of-thought
prompting~\cite{wei2022chainofthought}; single agent with
chain-of-inference; and a sequential multi-agent
decomposition~\cite{wu2023autogen}) on $30$ UK traffic signs
under $7$ visibility levels and $2$ occlusion modes, the two
judges are moderately associated ($r=0.644$). The asymmetric disagreement
--- $8.0\%$ of evaluations in which the $0$--$10$ judge scores
above $7$ while the equivalence judge rejects, peaking at
$14.2\%$ at visibility $v=0.8$ --- is the signal the dual-judge
protocol surfaces. A $49$-row human-eval validation confirms
that the $0$--$10$ judge tracks human judgement at $r=0.81$ and
the LLM accuracy sub-score at $r=0.80$, while the equivalence
judge applies a stricter wording-match criterion.
\paragraph{Contributions.} We make four contributions:
\begin{itemize}
\item \textbf{A controlled benchmark for vision-grounded regulatory
interpretation under perceptual occlusion} (\S\ref{sec:exp1}).
$4{,}680$ evaluations across $30$ UK traffic signs, $7$
visibility levels, $2$ occlusion modes, and $4$ VLM-based systems
on Azure GPT-4o. We release the benchmark and the occluded variants
for reproducibility.
\item \textbf{A dual-judge evaluation protocol} (\S\ref{sec:exp1})
pairing the standard $0$--$10$ LLM-as-judge with a strict binary
semantic-equivalence judge against human-curated gold
descriptions. The protocol adds one LLM call per evaluation. We
release the equivalence-judge prompt template so future legal-AI
work can adopt or extend it.
\item \textbf{A $49$-row human-eval validation}
(\S\ref{par:human-eval}) showing that both LLM judges track human
judgement (Pearson $r=0.81$ with the $0$--$10$ judge; $r=0.80$
with the LLM accuracy sub-score) and that the equivalence judge
applies a stricter---and, as we show, judge-dependent---reference-match
criterion.
\item \textbf{An empirical characterisation of dual-judge
disagreement}: its size ($8.0\%$ Type~II), its distribution across
visibility (reported \emph{conditioned} on the base rate, not only
marginally), and its judge-dependence. We discuss what this signal
contributes to existing AI-for-Law benchmark design in
Section~\ref{sec:discussion}.
\end{itemize}
\paragraph{A note on methodology.}
An earlier version of this work used a single LLM-as-judge to
determine whether a free-form description matched the ground-truth
sign type. A careful audit (Section~\ref{sec:audit}) revealed a
label-matching artefact that we address with a structured-output
fix. The dual-judge protocol builds on top of that corrected
pipeline.
\section{Related Work}
\paragraph{Multi-agent and chain-of-thought reasoning.}
A range of frameworks --- AutoGen~\cite{wu2023autogen},
multi-agent debate~\cite{du2024debate}, and other
patterns surveyed by~\citet{tran2025multiagent} --- propose
specialised LLM agents collaborating to outperform single-agent
prompting. Reported gains are mixed: most consistent on tasks
where sub-task decomposition is genuinely orthogonal, less so where
the integration step is the bottleneck. Chain-of-thought
prompting~\cite{wei2022chainofthought} elicits intermediate
reasoning from a single model and has been extended in many
directions, including zero-shot CoT~\cite{kojima2022zeroshot},
self-consistency decoding~\cite{wang2023selfconsistency},
Tree-of-Thoughts~\cite{yao2023tot}, and
ReAct~\cite{yao2023react}. In this paper we use four of these
systems as varied test subjects for our evaluation protocol; we
do not claim a new architectural contribution.
\paragraph{Traffic-sign recognition.}
Traffic-sign datasets such as the German Traffic Sign Detection
Benchmark~\cite{houben2013gtsdb} have driven decades of work on classical
detection and recognition pipelines. The shift to large-scale
vision-language models such as CLIP~\cite{radford2021clip} and
LLaVA~\cite{liu2023llava} has reopened the question of whether
zero-shot or instruction-tuned VLMs can directly interpret traffic
signs at semantic granularity (e.g., ``no stopping between 8am and
8pm'') rather than only at object-detection granularity.
Most VLM evaluations to date
use clean, full-visibility benchmarks; occlusion has been studied
primarily through standard object-detection metrics, and contextual
applicability (which sign applies to which lane) is typically assumed
away. We extend these evaluations by introducing controlled occlusion
for VLM-based systems.
\paragraph{LLM-as-judge evaluation.}
The use of strong LLMs to score the outputs of weaker
LLMs~\cite{zheng2023judging} has become standard practice but is
vulnerable to systematic biases (position, verbosity,
self-enhancement). Our methodology audit
(Section~\ref{sec:audit}) documents a related label-matching bias
against verbose systems and proposes a simple structured-output
fix.
\paragraph{Legal-AI benchmarks and evaluation.}
A growing line of work proposes benchmarks for AI systems applied
to legal tasks. \citet{chalkidis2022lexglue} introduce LexGLUE, a
multi-task benchmark covering legal-language understanding (case
classification, contract clauses, statutory citation) across seven
sub-tasks; \citet{guha2023legalbench} introduce LegalBench, a
collaboratively built benchmark of $162$ tasks spanning six types
of legal reasoning. Both benchmarks make essential progress in
standardising evaluation across legal NLP; in their primary
protocols, however, scoring is accuracy- or label-match-based,
which evaluates \emph{whether} a system produces a correct answer
rather than \emph{whether its response would survive a strict
audit against the reference}. Our paper contributes a small
empirical data point to that conversation. We
return to the implications for benchmark design in
Section~\ref{sec:discussion}.
\section{Methodology Audit and Metric Correction}
\label{sec:audit}
Our initial pipeline used a single LLM-as-judge~\cite{zheng2023judging}
to evaluate both descriptive quality (on a $0$--$10$ scale) and a binary
\texttt{correctly\_identified} flag. An audit revealed two systematic
inconsistencies: $4.8\%$ of evaluations had $\texttt{overall}\geq 7$
but $\texttt{correctly\_identified}=0$, concentrated in systems with
longer outputs (Multi-Agent, Chain-of-Inference); and ranking by judge
score gave $\text{CoT}>\text{CoI}>\text{SA}>\text{MA}$ while the
binary flag gave $\text{SA}>\text{CoI}>\text{CoT}>\text{MA}$, placing
the most thorough reasoner third on identification --- an implausible
capability ordering.
The diagnosis was that the judge was fuzzy-matching free-form
descriptions against ground-truth labels, so terser outputs
accidentally produced strings the judge parsed as the correct label
more reliably than longer ones. This is related to but distinct from
the verbosity bias of~\cite{zheng2023judging} (where longer responses
are \emph{over}-scored): here the penalty arises because the judge is
parsing labels from free text rather than scoring quality.
\textbf{Correction.} We required every system response to end with
a mandatory \texttt{SIGN\_TYPE: <category>} line drawn from a fixed
vocabulary of $25$ regulatory categories; sign-type identification
is now computed by a deterministic regex. Format compliance is
$99.5$--$100\%$.
The fix removes the parsing artefact; any residual rank differences
across the two judges (Table~\ref{tab:exp1_ranking}) then reflect the
genuine quality-versus-equivalence distinction this paper studies, not
a parsing defect. All results in subsequent sections use the corrected
pipeline.
\section{Experiment: Recognition Under Occlusion}
\label{sec:exp1}
\subsection{Setup}
\paragraph{Dataset.} We use a subset of $30$ UK traffic signs (TSRGD
codes spanning regulatory, warning, and direction categories). Our
choice of the UK regulatory vocabulary follows the convention used in
prior VLM-based traffic-sign work;
classical detection benchmarks such as
GTSDB~\cite{houben2013gtsdb} use German signs and a different
evaluation protocol focused on bounding-box detection rather than
semantic identification.
For each base sign, we generate occluded variants at seven visibility
levels ($100\%$, $80\%$, $70\%$, $50\%$, $30\%$, $20\%$, $10\%$)
under two occlusion modes:
\begin{itemize}
\item \textbf{Random-block:} a single opaque rectangle of area
$(1 - v)$ placed at a uniformly random position, simulating an
adjacent vehicle, tree, or pedestrian.
\item \textbf{Bottom-up:} a rectangle of height $(1 - v)$ growing from
the lower edge, simulating progressive occlusion from a vehicle directly
in front or a poor-lighting cutoff.
\end{itemize}
\begin{figure*}[t]
\centering
\includegraphics[width=\textwidth]{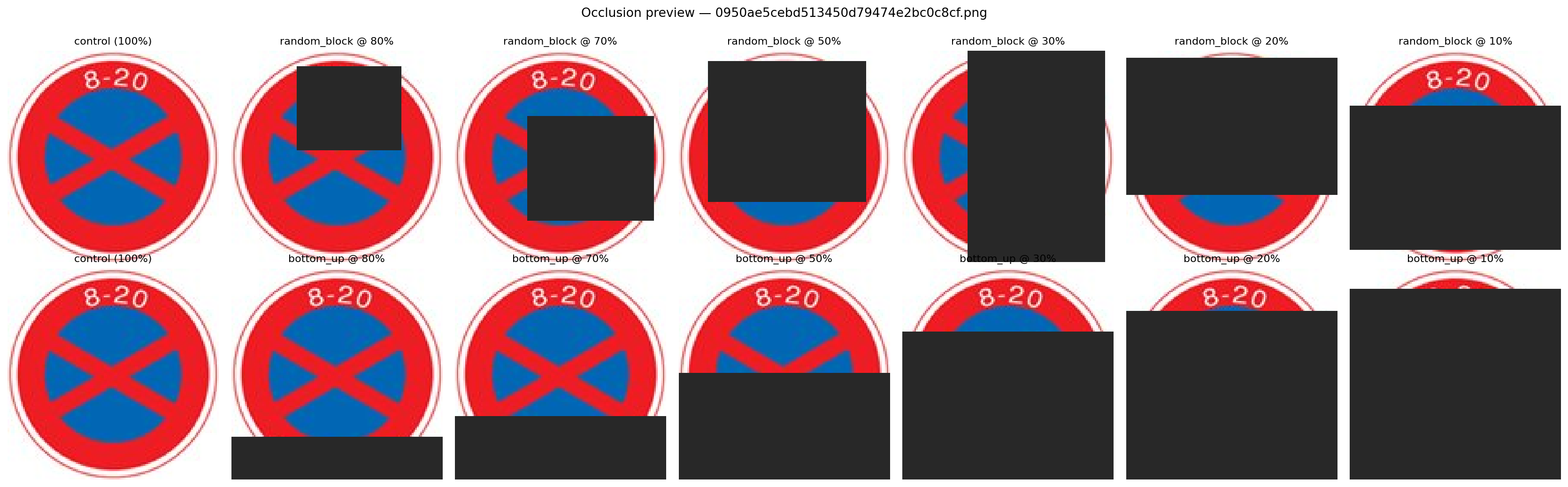}
\caption{The two occlusion modes applied to a representative
``no stopping'' regulatory sign across the seven visibility levels
($100\%$, $80\%$, $70\%$, $50\%$, $30\%$, $20\%$, $10\%$) used in our
quantitative experiments (Section~\ref{sec:exp1_results}).
\textit{Top row:} \textbf{random-block} occlusion, in which a
rectangle of area $(1-v)$ is placed at a uniformly random position,
simulating an adjacent vehicle, tree, or pedestrian.
\textit{Bottom row:} \textbf{bottom-up} occlusion, in which a
rectangle of height $(1-v)$ grows from the lower edge, simulating
progressive occlusion from a vehicle directly in front or a
poor-lighting cutoff. The leftmost column ($v=1.0$) is the
un-occluded control shared by both modes. The two modes degrade
the image very differently: bottom-up preserves the sign's upper
text region (``8--20'') down to roughly $v=0.30$, whereas
random-block can remove any part of the sign with equal
probability and at $v=0.20$ may obliterate the sign almost
entirely.}
\label{fig:occlusion_modes}
\end{figure*}
This yields $390$ variants ($30 \times 13$; the $v=1.0$ control is shared
across modes). Each variant is evaluated by every system under
$3$ independent runs, giving $4{,}680$ evaluations total.
\paragraph{Systems.} We evaluate four systems, all built on the same
Azure GPT-4o backbone~\cite{openai2024gpt4o}:
\begin{itemize}
\item \textbf{Single Agent (SA):} a single vision-call producing a
free-form description with a mandatory \texttt{SIGN\_TYPE} line.
\item \textbf{Single Agent (CoT):} as SA, with explicit chain-of-thought
prompting~\cite{wei2022chainofthought} eliciting four free-form
reasoning steps.
\item \textbf{Single Agent (CoI):} a chain-of-inference variant that,
unlike CoT's free-form reasoning, requires three explicitly separated and
ordered stages before the \texttt{SIGN\_TYPE} line --- (i) \emph{visual}
analysis (shape, colour, symbols), (ii) \emph{content} analysis (the rule
the sign states), and (iii) \emph{contextual} analysis (how and where the
rule applies). CoI differs from CoT in \emph{structure}, not information:
CoT reasons freely, whereas CoI imposes a fixed
visual$\rightarrow$content$\rightarrow$context decomposition. CoI is our
own variant and is not claimed as a contribution.
\item \textbf{Multi-Agent (MA):} sequential decomposition into Vision
Specialist $\to$ Text/Symbol Specialist $\to$ Integration Specialist,
implemented as a directed pipeline following the spirit
of~\cite{wu2023autogen}.
\end{itemize}
\paragraph{Ground truth and dual-judge evaluation.}
Ground truth in our experiments comes from a human-curated reference
table: each of the 30 source signs has an associated short
\textit{Caption} (e.g., ``Closed to Vehicles'') and a longer
\textit{Description} (e.g., ``Road is closed to all vehicles (cars,
light vehicles, motorcycles, etc.)'') prepared in advance by two
authors familiar with the UK regulatory vocabulary.
We stress that this reference is the authors' \emph{paraphrase} of each
sign's meaning, not the statutory text of the TSRGD or the Road Traffic
Act; grounding the reference in the governing instruments is future work.
We then apply \textit{two} LLM-as-judge protocols to every
system response. The first, anchored to the gold Description on three
$0$--$10$ axes (accuracy, completeness, relevance) and aggregated into
an \texttt{overall} score, is the same scoring style used in our
earlier pipeline~\cite{zheng2023judging}. The second is a strict
binary \textit{semantic-equivalence} judge that returns
$\textsc{equivalent}\in\{0,1\}$, asking only whether the predicted
description is semantically equivalent to the gold standard
(prompt template included in the supplementary material).
We report both judges, and analyse their agreement in
Section~\ref{sec:exp1_results} as a methodology-validation step.
\subsection{Results}
\label{sec:exp1_results}
\begin{figure*}[t]
\centering
\includegraphics[width=\textwidth]{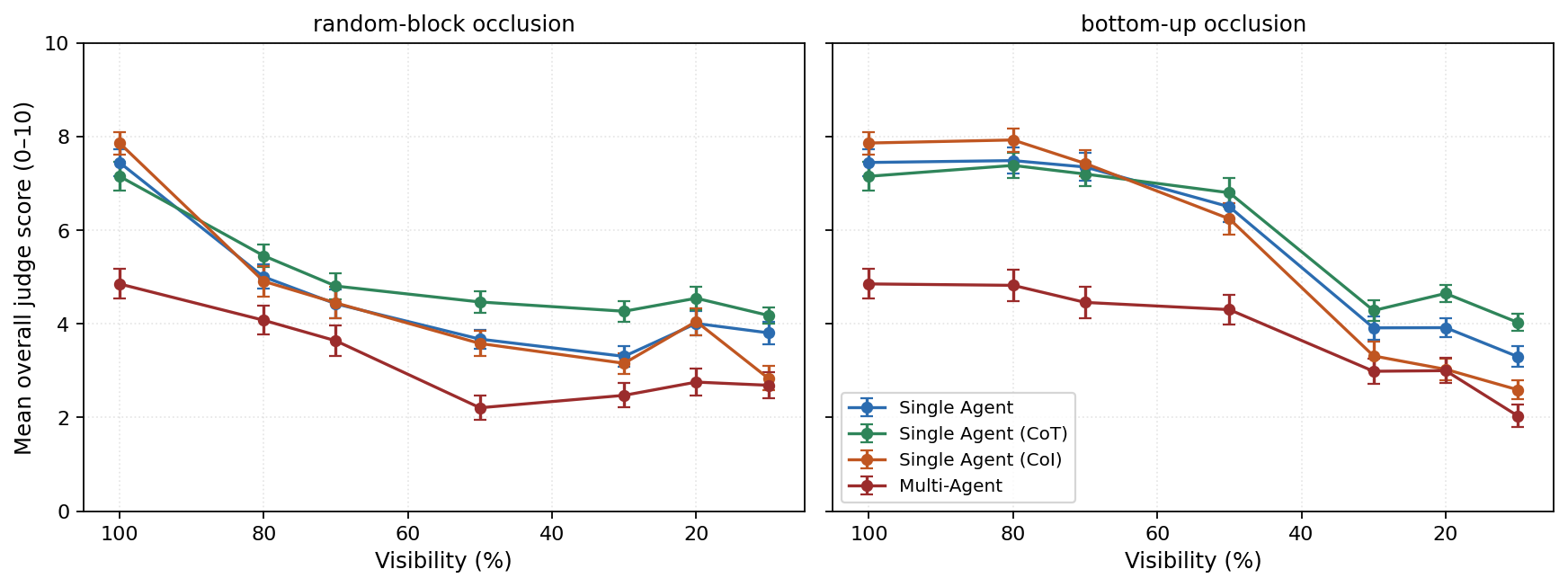}
\caption{Mean overall judge score ($0$--$10$, with error bars) under
the two occlusion modes, across all seven visibility levels.
Single-agent CoT (green) is consistently at or above the other
single-agent variants, while Multi-Agent (red) is dominated at every
visibility level. Degradation is smooth and monotonic under
\textbf{bottom-up} occlusion; under \textbf{random-block} the
degradation curve is steeper between $v=1.0$ and $v=0.5$,
reflecting that a single random rectangle of large area can quickly
hide most of the sign's discriminative content.}
\label{fig:exp1_curves}
\end{figure*}
\begin{figure}[t]
\centering
\includegraphics[width=\columnwidth]{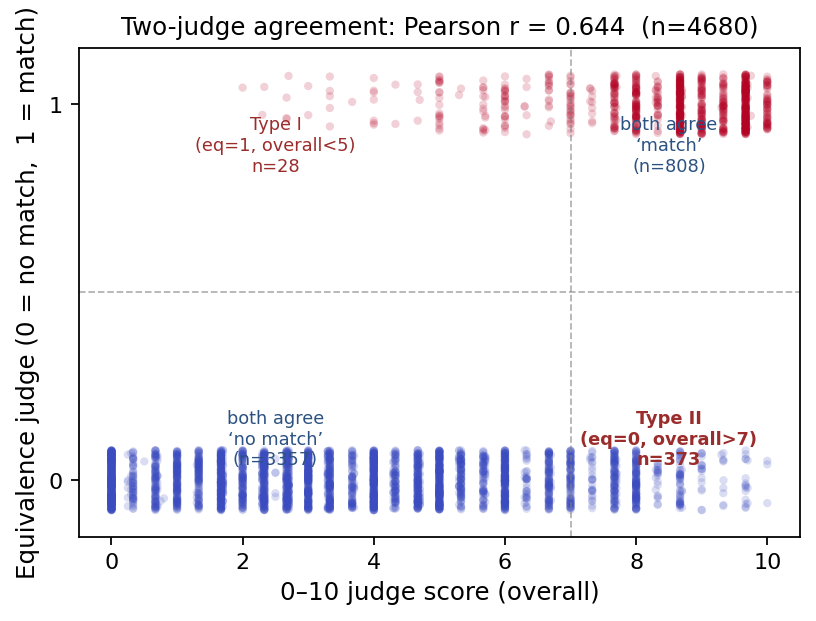}
\caption{Joint distribution of the two judges over all $4{,}680$
evaluations (vertical axis jittered). Point-biserial $r=0.644$.
The two off-diagonal quadrants hold the small Type~I set (the $0$--$10$
judge under-scored a semantically-equivalent response, $28$ rows) and
the substantially larger Type~II set (the $0$--$10$ judge over-scored a
response that failed strict semantic equivalence, $373$ rows); the
remaining $4{,}279$ rows ($91.4\%$) lie outside both disagreement
quadrants.}
\label{fig:exp1_twojudge}
\end{figure}
\begin{figure}[t]
\centering
\includegraphics[width=\columnwidth]{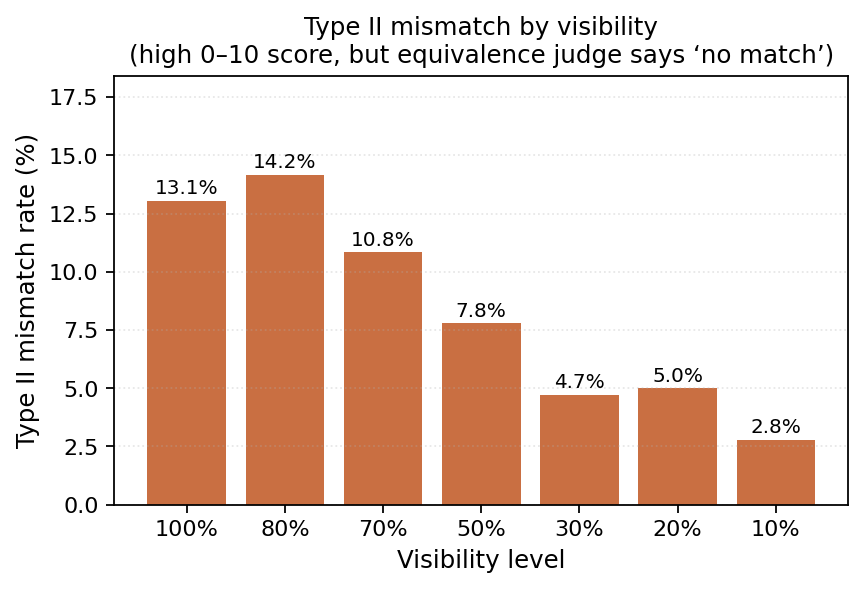}
\caption{Type II mismatch rate (\% of evaluations with
$\textsc{equivalent}=0$ but $\texttt{overall}>7$) at each visibility
level. The rate is highest at $v\in\{1.0, 0.8\}$ --- the clean-input
regime in which one would expect an LLM-as-judge to be most reliable ---
and broadly declines (though not strictly monotonically) as the image is
degraded further.}
\label{fig:exp1_type2}
\end{figure}

\begin{table}[t]
\caption{System means under the two judges, averaged across all
$4{,}680$ evaluations (raw aggregate means). \texttt{Overall} is the
$0$--$10$ judge anchored to the gold Description; \texttt{Equiv.\ rate}
is the fraction of responses the binary equivalence judge marks as
equivalent. Note the \emph{rank dissociation across judges}: CoT leads
on \texttt{Overall}, whereas CoI leads on \texttt{Equiv.\ rate} (and CoT
$0.234$ vs.\ SA $0.221$ are close but not tied). This dissociation is
itself an instance of the quality-versus-equivalence distinction this
paper studies, not a consistency check.}
\label{tab:exp1_ranking}
\begin{center}
\begin{small}
\begin{sc}
\begin{tabular}{lcc}
\toprule
System          & Overall ($0$--$10$) & Equiv.\ rate \\
\midrule
SA (CoT)       & $\mathbf{5.33}$ & $0.234$ \\
Single Agent   & $4.94$ & $0.221$ \\
SA (CoI)       & $4.72$ & $\mathbf{0.279}$ \\
Multi-Agent    & $3.41$ & $0.078$ \\
\bottomrule
\end{tabular}
\end{sc}
\end{small}
\end{center}
\end{table}
\begin{table}[t]
\caption{Paired $t$-tests on overall judge score against
Single Agent, computed per (source image, visibility level, mode)
cell after collapsing $3$ independent runs ($n=390$). Cells within a
sign are not fully independent across its $13$ variants (see text).}
\label{tab:exp1_ttest}
\begin{center}
\begin{small}
\begin{sc}
\begin{tabular}{lccc}
\toprule
System vs.\ SA & $\Delta_{\text{overall}}$ & $t$ & $p$ \\
\midrule
SA (CoT)    & $+0.391$ & $+4.89$  & $\mathbf{1.4\!\times\!10^{-6}}$ \\
SA (CoI)    & $-0.213$ & $-2.16$  & $0.032$ \\
Multi-Agent & $-1.527$ & $-12.38$ & $\mathbf{< 10^{-29}}$ \\
\bottomrule
\end{tabular}
\end{sc}
\end{small}
\end{center}
\end{table}
\paragraph{Headline finding.} Single-agent CoT significantly
outperforms the Single Agent baseline
(paired $\Delta=+0.39$, $p<10^{-5}$), while Multi-Agent significantly
\emph{underperforms} it ($\Delta=-1.53$, $p<10^{-29}$); CoI is
marginally below baseline ($\Delta=-0.21$, $p=0.032$). The
dominance of CoT and the underperformance of Multi-Agent are robust
under the $n=390$ paired analysis, though we note the $390$ cells are
not fully independent across the $13$ variants of a sign; sign-level
clustering would be the fully rigorous treatment and the two large
effects survive it.
\paragraph{Degradation behaviour.}
Figure~\ref{fig:exp1_curves} shows the per-system degradation across
all seven visibility levels under both occlusion modes. The ranking
is stable across the entire degradation range: Multi-Agent remains
dominated even at $v=0.1$ under random-block. Bottom-up degrades
more gradually because the sign's upper region (where discriminative
text and symbols sit) is preserved down to moderate visibility;
random-block can hide any region with equal probability and so its
degradation curve is steeper between $v=1.0$ and $v=0.5$.
\paragraph{Two-judge association (methodology validation).}
\label{par:two-judge-agreement}
To probe whether the $0$--$10$ judge tracks identification quality
against the gold reference, we compare it to the binary
equivalence judge over all $4{,}680$ evaluations. The two are
moderately associated (point-biserial $r=0.644$). We report this
association descriptively rather than as a hypothesis test: because the
rows are nested (three runs within each of $13$ variants within each
sign), a $p$-value computed as if the rows were independent is not
meaningful, so we do not report one; a sign-clustered or mixed-effects
estimate is the appropriate inferential treatment. Mean
$\texttt{overall}$ given $\textsc{equivalent}=1$ is $8.44$
(s.d.\ $1.44$); given $\textsc{equivalent}=0$ it is $3.62$
(s.d.\ $2.48$). The off-diagonals
(Figure~\ref{fig:exp1_twojudge}) are not symmetric: only $28$
rows ($0.6\%$; Type I) have $\textsc{equivalent}=1$ with
$\texttt{overall}<5$, but $373$ rows ($8.0\%$; Type II) have
$\textsc{equivalent}=0$ with $\texttt{overall}>7$ --- responses
that read well to the $0$--$10$ scale yet fail strict semantic
equivalence against the reference.
\paragraph{Where Type~II concentrates, and conditioning on the base rate.}
Marginally, the Type~II share of \emph{all} evaluations at each
visibility level is $13.1\%$ ($v=1.0$), $14.2\%$ ($v=0.8$), $10.8\%$,
$7.8\%$, $4.7\%$, $5.0\%$, and $2.8\%$ ($v=0.1$): highest where one would
naively expect the LLM-as-judge to be most reliable, and broadly
declining --- though \emph{not strictly monotonically}, since it rises
from $v=1.0$ to $v=0.8$ and again from $v=0.3$ to $v=0.2$
(Figure~\ref{fig:exp1_type2}). This marginal pattern is partly
mechanical: quality scores fall with occlusion, so fewer answers score
above $7$ at low visibility and there is simply less room for Type~II.
To separate the effect from this floor, Table~\ref{tab:cond} reports the
\emph{conditional} Type~II rate --- the share of \emph{high-scoring}
answers ($\texttt{overall}>7$) that the equivalence judge rejects ---
alongside the count of high-scoring answers per level. Conditioning
\emph{reverses} the marginal picture. The marginal rate peaks at high
visibility only because high-scoring answers are common there
($N_{>7}=217$--$319$ at $v\geq0.5$); \emph{conditionally}, a high quality
score is \emph{least} trustworthy under heavy occlusion, where Type~II
reaches $54$--$63\%$ of the few high-scoring answers that remain
($N_{>7}=37$--$66$ at $v\leq0.3$), versus $22$--$32\%$ at high visibility.
The correct reading is therefore not that disagreement concentrates on
clean inputs, but that a high quality score \emph{on a degraded input}
is the least reliable: when the sign is barely visible yet the quality
judge still scores a response above $7$, more often than not it fails
strict equivalence.
\begin{table}[t]
\caption{Type~II conditioned on the base rate. $N_{>7}$ is the number of
answers with $\texttt{overall}>7$ at each visibility level; the
conditional rate is (Type~II count)$/N_{>7}$. Reporting the conditional
rate separates the high-visibility concentration from a floor effect.}
\label{tab:cond}
\begin{center}
\begin{small}
\begin{sc}
\begin{tabular}{lccc}
\toprule
Visibility & $N_{>7}$ & Type~II count & Cond.\ rate \\
\midrule
$1.0$ & $217$ & $47$  & $21.7\%$ \\
$0.8$ & $319$ & $102$ & $32.0\%$ \\
$0.7$ & $283$ & $78$  & $27.6\%$ \\
$0.5$ & $205$ & $56$  & $27.3\%$ \\
$0.3$ & $54$  & $34$  & $\mathbf{63.0\%}$ \\
$0.2$ & $66$  & $36$  & $54.5\%$ \\
$0.1$ & $37$  & $20$  & $54.1\%$ \\
\bottomrule
\end{tabular}
\end{sc}
\end{small}
\end{center}
\end{table}
The interpretive consequence is direct, and it is about \emph{trust in a
high score}: at full visibility a response scoring above $7$ fails strict
equivalence about one time in five ($21.7\%$), whereas at heavy occlusion
such a response fails more often than not ($54$--$63\%$). A high quality
score should therefore be discounted most, not least, when the input is
degraded. We discuss the implications for AI-for-Law evaluation
methodology in Section~\ref{sec:discussion}.
\paragraph{Human-eval validation on a stratified sample.}
\label{par:human-eval}
The two-judge agreement reported above is between two LLM-based
judges (both GPT-4o, asked different questions about the same
response). To check whether either judge tracks human judgement, the
first author manually annotated a stratified sample of $49$
evaluations from the $4{,}680$-row dataset:
one row per (visibility-level, system) cell ($28$ rows),
plus $12$ Type II cases ($\textsc{equivalent}=0$ but
$\texttt{overall}>7$, the cells that drive the
methodology-validation argument), plus $9$ ``agreed-good'' cases
($\textsc{equivalent}=1$ and $\texttt{overall}\geq 7$).
\paragraph{Annotation protocol.}
For each row, the annotator was shown the gold caption, the gold
description, the model response, and the (visibility, mode, system)
metadata, but \emph{not} the LLM judge scores. Two judgements were
elicited per row using a fixed rubric (full version in the
supplementary material): a binary semantic-equivalence label
($1$ if the response identifies the same sign as the gold reference
despite wording differences; $0$ if it misidentifies the sign,
gives a generic appearance description, refuses, or returns
\texttt{SIGN\_TYPE: other}); and a $0$--$10$ quality score anchored
at $9$--$10$ (perfect identification), $7$--$8$ (correct with minor
gap), $5$--$6$ (general category right), $3$--$4$ (wrong but
reasonable given visibility), $1$--$2$ (vague or generic), and
$0$ (refused or nonsensical). The annotator was instructed not to
grade on a curve, not to penalise verbosity, and to judge against
the gold rather than an imagined version of the sign.
\paragraph{Results.}
The two questions ask the same thing of the human as of the LLM
judges.
\begin{itemize}
\itemsep -0.2em
\item \textbf{$0$--$10$ score:} Pearson $r=0.81$ between
human and the aggregated $\texttt{overall}$ judge
($p<10^{-12}$, $n=49$); Spearman $\rho=0.73$ ($p<10^{-8}$).
Human and LLM means are within $0.14$ points of each other
(human $\overline{x}=6.43$, LLM $\overline{x}=6.29$). The $0$--$10$
judge tracks human judgement closely.
\item \textbf{LLM-accuracy sub-score:} The $0$--$10$ judge returns
three sub-scores (accuracy, completeness, relevance) that are
aggregated into $\texttt{overall}$. Because the accuracy sub-score
is the one most directly comparable to a human ``did the system
get the sign right'' judgement, we report it separately:
$\textsc{Pearson}$ between $\texttt{human\_overall}$ and the LLM
accuracy sub-score is $r=0.80$, $p<10^{-11}$, $n=49$.
The three sub-scores correlate with human judgement at
$r \in [0.79, 0.80]$, indicating that no single sub-score is
driving the agreement: the $0$--$10$ judge tracks humans on
\emph{accuracy}, completeness, and relevance separately.
\item \textbf{Binary equivalence:} Cohen's $\kappa=0.38$~\cite{cohen1960kappa}, raw
agreement $65\%$ ($32$ of $49$). The disagreement is entirely in one
direction: the LLM never marks ``match'' when the human does not,
while it marks ``no match'' for $17$ responses the human marked
``match'' (these $17$ have human $\overline{\texttt{overall}}=7.06$,
$\textrm{sd}=1.68$). A $\kappa$ of $0.38$ indicates fair (not
strong) agreement; we interpret this asymmetric disagreement as
informative rather than as a flaw, but a multi-annotator study is
needed to disentangle annotator-specific patterns from systematic
differences between human and equivalence-judge criteria.
\end{itemize}
The asymmetric disagreement is methodologically informative: the
equivalence judge rejects responses that get the general category
right but do not match the reference's specific wording. For
example, when the gold says \emph{``you must slow down to a level
where you can stop immediately''} and the system says
\emph{``warning to drivers to reduce speed and proceed with
caution''}, a human treats these as equivalent in everyday meaning
but a legal auditor would note the response omits the
``immediately stoppable'' criterion. The equivalence judge sides
with the auditor. The full annotation file is
in the supplementary material.
\section{Discussion}
\label{sec:discussion}
\paragraph{What the dual-judge protocol contributes.}
The two-judge agreement reported in
Section~\ref{par:two-judge-agreement} ($r=0.644$, $91.4\%$ of rows
outside the disagreement quadrants, $8.0\%$ Type II and $0.6\%$ Type I)
demonstrates a useful property of the protocol: most of the time,
two GPT-4o judges asked different questions about the same response
give compatible answers. The asymmetric disagreement is the
informative signal. The $0$--$10$ judge and the equivalence judge
do not over- and under-credit symmetrically: the disagreement is
one-directional, and (once we condition on the answer already scoring
highly; Table~\ref{tab:cond}) it is \emph{most} severe under heavy
occlusion, where $54$--$63\%$ of high-scoring answers fail strict
equivalence. A visibility-dependent, one-directional disagreement signal
is what the dual-judge protocol contributes that a single quality score
does not.
\paragraph{Verbalisation, not interpretation.}
Our task is rule \emph{verbalisation} --- stating the rule a sign
encodes --- not rule \emph{application}, the open-textured question of
whether a rule governs a given set of facts. We chose traffic-sign
meaning precisely because it is closed and codified, with (in the
ordinary case) a single correct reading; much textual legal work lacks
this property, and there the equivalence question is itself contested and
multi-factor. We therefore do not claim the protocol evaluates legal
reasoning in general. Extending from verbalisation to applied-rule
outputs (e.g., ``given the current time, is stopping here permitted?'')
is a concrete and important next step.
\paragraph{The Type II signal interpreted for legal AI.}
For legally accountable deployment, the relevant question is
whether a response would survive an audit against the applicable
reference. The standard $0$--$10$ judge measures something
related but not identical: whether the response is fluent and
informative. The equivalence judge, by contrast, applies a strict
wording-match criterion.
Two claims must be kept separate. That the equivalence judge is
\emph{stricter} than the everyday reader is established by our data: a
merely noisy judge would disagree in both directions, whereas this one
errs only toward rejection. That its higher bar is \emph{legally}
grounded --- strict in the direction a statute would require, rather than
simply more conservative --- is \emph{not} established, since the
reference is our own paraphrase rather than statutory text and no legally
trained annotator was involved. We therefore use ``audit-grade'' as a
motivating analogy, not a validated property, and read the Type~II
quantity as a property of this judge and this reference. Our human-eval
validation (Section~\ref{par:human-eval}) clarifies which view is which:
humans and the $0$--$10$ judge align with everyday-reader judgement
($r=0.81$); the equivalence judge is fairly but one-directionally
stricter.
\paragraph{How this complements LexGLUE and LegalBench.}
LexGLUE~\cite{chalkidis2022lexglue} and
LegalBench~\cite{guha2023legalbench} have driven essential
standardisation across legal NLP and legal-reasoning evaluation,
covering tasks where a gold label or expert adjudication is
available. The dual-judge protocol we propose is complementary: it
applies to open-ended responses where a gold reference exists
but exact label-match is not the natural protocol, and adds one
semantic-equivalence call per evaluation. The protocol does not
replace accuracy-style scoring; it provides one additional
signal --- the equivalence-judge rejection rate --- alongside it.
We release our equivalence-judge prompt template in the
supplementary material so other benchmarks can adopt the
same signal at low cost.
\paragraph{Scope of generalisation.}
The dual-judge \emph{mechanism} is modality-independent and would
transfer to text without difficulty, since it concerns only how an
output is scored against a reference. The empirical \emph{findings},
by contrast, are tied to properties of the visual testbed: the
difficulty gradient is produced by occlusion, for which textual law
has no clean analogue, and the tractability of the task is purchased
by a closed, codified reference. We therefore scope our empirical
claims to visually grounded regulatory tasks --- such as the deployed
driving and enforcement systems this domain motivates --- and not to
textual legal reasoning in general. We do not claim that the $8\%$
figure transfers; different domains will have different disagreement
rates.
\paragraph{Shared-model bias.}
Both judges and all four evaluated systems are GPT-4o, exposing the
judges to a documented self-preference effect. We disclose this as a
genuine limitation: before the disagreement signal is treated as
model-independent, a sample of equivalence verdicts should be
corroborated by a judge from a different model family. We did not do so
here and flag it as required future work.
\paragraph{Methodology audit as a complementary contribution.}
The structured-output fix in Section~\ref{sec:audit} is logically
independent of the dual-judge protocol, but the two work well
together. The audit removes a known parsing artefact: a single
LLM-as-judge asked to fuzzy-match free-form descriptions against
ground-truth labels can over-credit terser outputs whose surface
form happens to match. Requiring structured output (a mandatory
\texttt{SIGN\_TYPE} line parsed with a deterministic regex) closes
that gap. The dual-judge protocol then surfaces \emph{residual}
disagreement between ``fluent and informative'' and
``matches the reference'' that no single judge can capture by
construction. Both steps are easy to adopt: the structured-output
fix takes one prompt edit and one regex; the dual-judge protocol
takes one additional LLM call per evaluation.
\paragraph{Limitations.}
Our dataset is small ($30$ base signs), and the experiments use a
single VLM backbone~\cite{openai2024gpt4o}; extending to GPT-5 and
open VLMs is future work. Both LLM judges are GPT-4o instances
asked different questions, so some of the agreement we report may
reflect shared judge biases rather than independent agreement; the
$49$-row human-eval study supports the LLM judges' calibration but
is a single-annotator sanity check, and a multi-annotator
extension with at least one legally trained annotator is the
natural next step. We also note that the annotator was the first
author, who wrote the rubric and is invested in the paper's
claims; while annotation was conducted without sight of the LLM
judge scores, ideal validation requires independent annotators.
The reference is an author paraphrase rather than statutory text.
Our evaluation does not include the German
GTSDB benchmark~\cite{houben2013gtsdb}, which is detection-focused
under a different protocol.
\section{Conclusion}
We have presented a dual-judge evaluation protocol for vision-language
models on a visually grounded regulatory task, together with a controlled
testbed and a small human-eval validation. The protocol pairs the
standard $0$--$10$ LLM-as-judge with a strict binary
semantic-equivalence judge against human-curated gold descriptions,
adding one additional LLM call per evaluation. Across $4{,}680$
traffic-sign evaluations under seven visibility levels and two
occlusion modes, the two judges are moderately associated (Pearson
$r=0.644$) with an asymmetric $8\%$ disagreement signal whose marginal rate peaks at
high visibility ($14.2\%$ at $v=0.8$) but whose \emph{conditional} rate is
highest under heavy occlusion ($54$--$63\%$ of high-scoring answers at
$v\leq0.3$), so a high quality score is least trustworthy when the input is
most degraded.
A $49$-row human-eval validation confirms that the $0$--$10$ judge
aligns with everyday-reader judgement (Pearson $r=0.81$) and that
the equivalence judge is fairly but one-directionally stricter.
For the AI-for-Law community, the practical contribution is a
low-cost evaluation addition: a single extra LLM call per response
surfaces a visibility-dependent disagreement signal that
single-judge protocols, by construction, do not report. The
protocol complements rather than replaces existing legal benchmarks
such as LexGLUE~\cite{chalkidis2022lexglue} and
LegalBench~\cite{guha2023legalbench}. We frame the equivalence
signal as judge-dependent and the legal language as motivation, and we
scope the empirical claims to visually grounded regulatory tasks. We
release the equivalence-judge prompt template, the occluded variants,
and the full evaluation results in the supplementary material so that
future evaluation work can adopt, critique, or extend the protocol.
\section*{Software and Data}
The prompt templates, occluded variants, and full evaluation results
are released at \url{https://github.com/ImSuMyatNoe/dual-judge-traffic-signs}.
\section*{Impact Statement}
This paper presents work whose goal is to advance the evaluation of
AI systems intended for legally accountable deployment. We argue
that benchmarks in this area should report a strict equivalence signal
alongside a quality score, on the grounds that the two can dissociate in
practice. We caution that the equivalence judge is itself only fairly
validated and should not be treated as an authoritative legal standard
without further, independent validation. We do not see specific ethical
concerns beyond those generally applicable to work on autonomous systems
and AI evaluation.
\section*{Acknowledgements}
This work was supported by the ``Strategic Research Projects" grant from ROIS (Research Organization of Information and Systems), the ``R\&D Hub Aimed at Ensuring Transparency and Reliability of Generative AI Models" project of the MEXT, by JSPS KAKENHI Grant Numbers, 25H00522 and 25H01112, and JST as part of Adopting Sustainable Partnerships for Innovative Research Ecosystem (ASPIRE), Grant Number JPMJAP25B2.
\bibliography{example_paper}
\bibliographystyle{icml2026}
\end{document}